\documentclass[conference]{IEEEtran}
\IEEEoverridecommandlockouts
\IEEEoverridecommandlockouts

\usepackage{cite}
\usepackage{amsmath,amssymb,amsfonts}
\usepackage{algorithmic}
\usepackage{graphicx}
\usepackage{textcomp}
\usepackage{xcolor}
\usepackage{booktabs}
\usepackage{multirow}
\usepackage{url}
\usepackage{stfloats}
\usepackage{flushend}

\def\BibTeX{{\rm B\kern-.05em{\sc i\kern-.025em b}\kern-.08em
    T\kern-.1667em\lower.7ex\hbox{E}\kern-.125emX}}

\begin{document}

\title{Cost-Aware Reinforcement Learning with Action\\
Masking and Projection for Battery Energy Storage\\
Dispatch under Suppressed-Spread Market Shifts\thanks{\copyright~2026 IEEE. Personal use of this material is permitted. Permission from IEEE must be obtained for all other uses, in any current or future media, including reprinting/republishing this material for advertising or promotional purposes, creating new collective works, for resale or redistribution to servers or lists, or reuse of any copyrighted component of this work in other works. Accepted for publication at IEEE IECON 2026.}}

\author{
\IEEEauthorblockN{Kuanlin Chen}
\IEEEauthorblockA{
\textit{Independent Researcher}\\
Taoyuan, Taiwan\\
dipper13579@gmail.com\\
ORCID: 0009-0006-3879-4565
}
\and
\IEEEauthorblockN{Chen-Wei Kuo}
\IEEEauthorblockA{
\textit{National Tsing Hua University}\\
Hsinchu, Taiwan\\
ck3294@nyu.edu
}
\and
\IEEEauthorblockN{Cheng-En Ou}
\IEEEauthorblockA{
\textit{Independent Researcher}\\
garyandevan3@gmail.com
}
}

\maketitle

\begin{abstract}
Battery energy storage system (BESS) dispatch must preserve operational
feasibility while declining price spreads reduce the margin available to pay
for cycling. We study a proximal policy optimization (PPO) controller whose
pre-selection physical action mask and emergency projection are separated from a causal,
forecast-informed economic advisory. All forecast-dependent methods receive the
same causal 24-step forecast and grid-side settlement. Across five PPO seeds,
advice-on net profit is 30.59 and 18.04 USD per 336-hour T1 and T2 window,
versus 36.77 and 22.94 USD for proxy-cost MPC; PPO remains below this
reference in both periods.
Advice raises T2 profit from 16.45 to 18.04 USD while reducing throughput, but
is immaterial in T1. On disjoint weekly blocks, PPO is stable under daily,
weekly, and blended seasonal forecasts, weakens under persistence, and remains
below proxy-cost MPC. Paired diagnostics localize changes to the observed
5--10 USD/MWh regime with mixed SoC-dependent effects. An M0--M6 ablation shows
that mask removal sends thousands of infeasible requests to projection, while
removing both physical layers exposes ramp violations. The evidence separates
economic screening from feasibility enforcement without claiming formal
safety, lifecycle-optimal aging, or RL dominance.
\end{abstract}


\section{Introduction}

Battery energy storage systems provide critical grid services by
arbitraging electricity prices, supporting renewable integration,
and maintaining grid stability.
However, real-time dispatch optimization is complicated by highly
volatile electricity prices, physical battery degradation
constraints, and short-horizon market opportunity shifts.

Model predictive control~\cite{mpc_ref} and rule-based
methods require forecasts and modelling assumptions whose performance can vary
when market conditions shift.
Deep reinforcement learning (DRL) has demonstrated promise in
sequential dispatch decision-making~\cite{rl_bess_ref,perez2026},
yet most prior work evaluates on in-distribution test sets,
obscuring brittleness under true distributional shift.

This paper makes the following contributions:

\begin{enumerate}
    \item \textbf{Causal cost-aware dispatch under suppressed spreads}: a
    common-information comparison of PPO and reference controllers under
    identical issued forecasts and grid-side settlement.

    \item \textbf{A layered dispatch interface}: a pre-selection physical
    action mask, downstream emergency projection, and a separate soft
    economic advisory based on a fixed throughput-cost proxy.

    \item \textbf{Mechanism-level evidence}: a five-seed M0--M6 ablation and
    same-state diagnostic attribute infeasible requests, projection, observed
    violations, and conditional action changes to individual layers.

    \item \textbf{Claim-bounded evidence}: common-accounting reference
    comparisons with explicit seed-level dispersion and claim limits.
\end{enumerate}

\section{Related Work}

\subsection{Deep RL for BESS Arbitrage}
PPO and SAC can be competitive in multi-market dispatch~\cite{ppo_ref,sac_ref};
utility-scale PV-BESS evidence also reports adaptive strategies~\cite{perez2026},
and MILP-guided DRL improves sample efficiency under forecast uncertainty
~\cite{milp_drl_ref}. DRL storage arbitrage has also been evaluated with an
explicit lithium-ion degradation model~\cite{cao2020drl}. These studies mainly use in-distribution tests and
simplified battery models, leaving shift robustness unresolved.

\subsection{Safe RL and Physical Constraints}
Action masking dynamically filters infeasible actions at the
policy-output level, giving a pre-selection feasibility mechanism that is
distinct from a penalty-only treatment and has direct policy-gradient
justification~\cite{invalid_action_masking}. Related power-system work uses
physics-shielded RL for voltage control~\cite{chen2023shield}, while safe-set
projection has been studied generally and for BESS imbalance
settlement~\cite{gros2020projection,rasic2025safe_bess}.
Arrhenius-based degradation has been integrated into RL reward
functions at one-second resolution for frequency regulation
markets~\cite{fcrn_ref}.
Constrained Policy Optimization~\cite{cpo_ref} extends
trust-region methods with explicit multi-dimensional constraint
guarantees.

\subsection{Market Shifts and Stress-Augmented Training}
The COVID-19 pandemic caused unprecedented market anomalies:
NYISO recorded wholesale prices as low as \$25.70/MWh with a
31\% decline in day-ahead congestion value~\cite{nyiso2020}.
DR-SPCRL addresses OOD brittleness by scheduling the
distributional robustness budget as a curriculum variable,
achieving 24.1\% higher episodic return under
perturbation~\cite{drspcrl_ref}.
Curriculum learning~\cite{curriculum_ref} improves
generalization by ordering training experiences from easy
to difficult. In this study, stress augmentation defines the trained
checkpoint configuration; no isolated curriculum-effect claim is made.

\section{Problem Formulation}

\subsection{MDP Formulation}
We formulate BESS dispatch as a Markov Decision Process
$\langle \mathcal{S}, \mathcal{A}, P, R, \gamma \rangle$
with hourly time steps.

\textbf{State space} $\mathcal{S} \in \mathbb{R}^{60}$:
12 scalar features (SoC, temperature, SoH, contemporaneous price,
net load, carbon intensity, PV/wind generation, temporal encodings,
and price volatility) and 48 forecast features. The 24-step price and
net-load forecasts are issued at decision time $t$ by a causal
seasonal-naive procedure using only the corresponding daily and weekly
historical lags; no realised value with target time after $t$ is read.

\textbf{Action space} $\mathcal{A}$: 21 discrete actions
$\{-1.0, -0.9, \ldots, 0, \ldots, +0.9, +1.0\}$ representing
charge/discharge power as a fraction of rated capacity
(2.5\,MW / 10\,MWh).

\textbf{Reward}:
\begin{equation}
r_t = p_t e_t + \lambda_\text{REC} g_t\max(e_t,0)
    - c_\text{proxy}|e_t|
    - \lambda_T \max(0,\, T_t - 45)
\end{equation}
where $p_t$ is the price for the delivery interval $[t,t{+}1)$,
$e_t$ is net grid-side export in that interval (positive for discharge), and
$g_t$ is the normalized clean-energy credit fraction. We use
$\lambda_\text{REC}=30$\,\$/MWh and $c_\text{proxy}=10$\,\$/MWh. The degradation term is a fixed throughput proxy
plus a logged Arrhenius state-change diagnostic; it is not interpreted as
a validated marginal aging cost. $\lambda_T = 100$\,\$/\,$^\circ$C penalizes thermal
excursions above 45\,$^\circ$C.
Note that 40\,$^\circ$C is the fan activation threshold
(passive-to-active cooling switch), not a hard constraint.

\section{Methodology}

\subsection{Reduced-Order Physics-Informed BESS Environment}

Unlike a lossless integrator, the environment includes
reduced-order coupled electrical, thermal, and state-of-health calculations
for feasibility, diagnostics, and derating. It is not presented as a
validated plant digital twin or as a lifecycle-optimization model.

\textbf{Thevenin Equivalent Circuit.}
Terminal voltage is computed via a first-order RC network:
\begin{equation}
V_t = \text{OCV}(\text{SoC}) - I_t R_0 - V_{RC,t}
\end{equation}
where $\text{OCV}(S) = 340 + 50(S{-}0.5) + 8\ln S
- 8\ln(1{-}S)$, $R_0$ varies with SoH and temperature,
and $V_{RC}$ captures polarization dynamics.
Voltage collapse protection automatically derates the power
setpoint when the discriminant $\Delta < 0$.

\textbf{Arrhenius Thermal Aging.}
SoH degradation per time step is:
\begin{equation}
\Delta\text{SoH} = K \cdot
\exp\!\left(\frac{-E_a}{R_g\,T}\right)
\cdot \frac{|I|}{C_n} \cdot \Delta t
\label{eq:arrhenius}
\end{equation}
with $E_a = 24{,}500$\,J/mol (within a literature-reported NMC
range of 22{,}500--28{,}775\,J/mol~\cite{liu2021,barletta2022}),
$R_g = 8.314$\,J/(mol$\cdot$K), $K = 2{\times}10^{-9}$,
and an adopted equivalent-pack energy capacity of 10\,MWh.
Dynamic derating activates when SoH $< 0.90$, limiting
maximum power to protect aged cells.
For the current-based diagnostic, $C_n$ is represented in Ah
(10\,MWh / 400\,V $\approx$ 25{,}000\,Ah), and $\Delta t$ is accumulated
via physics substeps, yielding an internally unit-consistent
C-rate$\cdot\Delta t$[h] diagnostic within the adopted equivalent-pack aggregation.

\begin{table}[t]
\centering
\caption{Battery Environment Parameters}
\label{tab:battery_params}
\begin{tabular}{lll}
\toprule
Parameter & Value & Basis \\
\midrule
Nominal energy capacity & 10 MWh & System design \\
Rated power & 2.5 MW & System design \\
$E_a$ (Arrhenius) & 24,500 J/mol & NMC range~\cite{liu2021,barletta2022} \\
$K$ (pre-exponential) & $2{\times}10^{-9}$ & Implementation setting \\
$R_0$ (internal resistance) & 0.0035 $\Omega$ & \cite{barletta2022} \\
EOL threshold (SoH) & 0.80 & Implementation setting \\
Derating start (SoH) & 0.90 & System design \\
$R_{th,passive}$ & 0.001 K/W & \cite{barletta2022} \\
$R_{th,active}$ & 0.0001 K/W & System design \\
Fan trigger temp. & 40 $^\circ$C & System design \\
$P_{cool}$ & 5 kW & System design \\
\bottomrule
\end{tabular}
\end{table}

\textbf{Dual-Mode Thermal Management.}
Cell heat combines Joule heating and entropic heat.
Passive cooling ($R_\text{th} = 0.001$\,K/W) operates below
40\,$^\circ$C; active fan cooling
($R_\text{th} = 0.0001$\,K/W) engages above this threshold,
drawing auxiliary power $P_\text{cool} = 5$\,kW.

\textbf{Carbon Diagnostic.}
The environment tracks battery carbon intensity as a charging-weighted moving
average over stored and charged energy, yielding the normalized credit
$g_t = \max(0,\,\text{CI}_\text{grid}-\text{CI}_\text{batt})/400$ as a
reward/observation component. The normalized factor enters (1) only on
discharge intervals through the $\max(e_t,0)$ term. Because no carbon objective is evaluated, this
credit serves only as an internal charging-source diagnostic and is not
interpreted as carbon-free-energy accounting or compliance~\cite{google_cfe}.

\textbf{Pre-selection Physical Action Mask.}
At each decision time, infeasible actions receive logit $-\infty$
before softmax.
The mask enforces SoC bounds (0.02--0.98), SoH-proportional
power derating (activates below SoH\,$<$\,0.90), and
ramp-rate limits ($\pm$0.2\,p.u./step).
Physical feasibility is kept separate from a forecast-informed economic
advisory. The advisory applies a soft action preference based on the issued
forecast and fixed throughput proxy; only physical infeasibility can cause
emergency projection of a requested action. Thermal and voltage conditions
are monitored by the implemented environment physics. Consequently, this
study reports constraints enforced or monitored under the evaluated scenarios,
rather than a hard safety guarantee beyond those scenarios.

\subsection{Training Configuration}

The selected checkpoint comes from a fixed three-phase schedule: historical
2018--2019 training, 5\% mild stress injection, and 10\% injection of price,
load, carbon, and compound perturbations calibrated only from the 2012--2017
design set. The 2020-January validation set selects a checkpoint by
$\min(V_\text{profit},V_\text{stress})$. Because the corrected study does not
isolate this schedule in a controlled training-time comparison, it is reported
as configuration rather than a contribution; the mechanism results instead
use evaluation-time toggles on the same selected checkpoints.

\subsection{PPO Actor Parameterization}

The implementation uses a shared recurrent feature encoder and discrete PPO
policy head. Architectural variants, including the legacy mixture-of-experts
configuration, are treated as implementation choices rather than as a central
empirical claim in this corrected study.

\subsection{PPO Training Details}

We use clipped PPO~\cite{ppo_ref} with:
value loss coefficient $= 0.5$,
KL early stopping ($\delta_\text{KL} = 0.02$),
entropy annealing ($0.02 \to 0.001$),
and online EMA reward normalization.
$K = 10$ epochs per update, batch size 1024,
learning rate $3{\times}10^{-4}$, $\gamma = 0.99$.

\section{Experimental Setup}

\subsection{Dataset and Temporal Isolation}

We use the NYISO electricity market dataset with co-located
solar, wind generation, and carbon intensity profiles.
The declared temporal split separates the training, validation, and test files:
\begin{itemize}
    \item \textbf{Train}: 2018--2019 historical (Phases 1--3)
    \item \textbf{Val}: 2020-Jan (checkpoint selection only)
    \item \textbf{T1}: Feb 1--Mar 21, 2020 (normal market)
    \item \textbf{T2}: Apr 1--Jun 7, 2020 (suppressed-spread test)
\end{itemize}
T2 boundaries correspond to the NY State PAUSE executive order
(Mar 22) and NYC Phase~1 reopening (Jun 8)---administrative
events determined prior to any data inspection~\cite{nyiso2020}.
The 24-step price and load forecasts are causal seasonal-naive forecasts
issued at each decision time. For each target time, the forecast combines
daily and weekly lagged values whose timestamps are no later than the issue
time. The PPO policy, economic advisory, and forecast-based MPC baselines
receive the same issued forecast bundle. Hence, results are interpreted as
a forecast-informed short-horizon dispatch setting; the forecast procedure
is a deployment limitation rather than a new forecasting contribution.

\subsection{Baselines}

\textbf{Price MPC (forecast)}: A price-only 24-hour MPC reference that uses
the same issued causal price forecast as the PPO controller. It maximizes
forecast trading revenue subject to the implemented feasibility interface.

\textbf{Proxy-Cost MPC (forecast)}: A 24-hour MPC reference that uses the same
forecast bundle and a fixed throughput penalty. It is a short-horizon economic
screening comparator, not a lifecycle-optimal controller.

\textbf{Rule-Based}: Threshold-based charge/discharge heuristic
with 7-day causal rolling window
(discharge: $\mu{+}2\sigma$, charge: $\mu{-}0.5\sigma$).
The rule parameters are fixed before test evaluation.

\textbf{PPO advisory on/off}: The same five selected PPO checkpoints are
evaluated greedily with only the economic-advisory toggle changed. This is the
paired controller comparison used to identify advisory effects.

The prior NoVirus training variant is retained as an auxiliary negative check,
but it is not a primary baseline: the corrected evidence does not isolate a
curriculum effect from training variation.

\subsection{Evaluation-Time Mechanism Ablation}

Seven variants reuse the same five checkpoints. M0 enables the mask, projection,
and advisory; M1 disables advice; M2 disables the physical mask; M3 disables
emergency projection; M4 keeps only the two physical layers; M5 keeps only the
advisory; and M6 is the bare policy. Requested, filtered, and executed actions
are logged separately. This design distinguishes an infeasible request caught
before execution from an observed post-execution violation.

For the advisory diagnostic, issued-forecast spread is binned using only values
available at the decision time. Realised margin is joined afterward solely to
form ex-post missed-opportunity and unprofitable-cycle proxies; it is never a
controller or advisory input.

For the conditional diagnostic, each M0 decision state is copied once. The
greedy advice-on and advice-off actions are each stepped for one interval from
that identical state, while only the M0 transition continues the reference
trajectory. The resulting paired local differences are grouped by the declared
spread bins, SoC ($<0.30$, 0.30--0.70, $>0.70$), and the advice-off candidate
action. They are not treated as a second full-horizon policy rollout.

\subsection{Evaluation Protocol}

Each seed uses 20 predefined, possibly overlapping 336-hour windows per period;
overlap is not treated as independent temporal evidence. Greedy evaluation uses
realized prices and loads only for current delivery and settlement. BestVal is
selected on Jan.~2020 validation by Maximin. The primary metric is mean window
net profit under common accounting; deterministic references have one
realization and PPO has five independent seeds. A sensitivity test reuses fixed
checkpoints with persistence, daily, weekly, and blended forecasts on disjoint
168-hour blocks (five T1, eight T2). Future realizations score error only after
issuance and never enter controller inputs.

\section{Results}
\label{sec:results}

\IfFileExists{generated/camera_ready_results.tex}{%

\subsection{Suppressed-Spread Market Context}
Figure~\ref{fig:market_regime} establishes the evaluated shift.  The mean
24-hour spread in T2 falls to 27.8\% of its training-period value.  This is a
short-horizon opportunity shift, not evidence that every T2 interval is less
volatile or that the controller observes future realised prices.

\begin{figure}[t]
\centering
\includegraphics[width=\columnwidth]{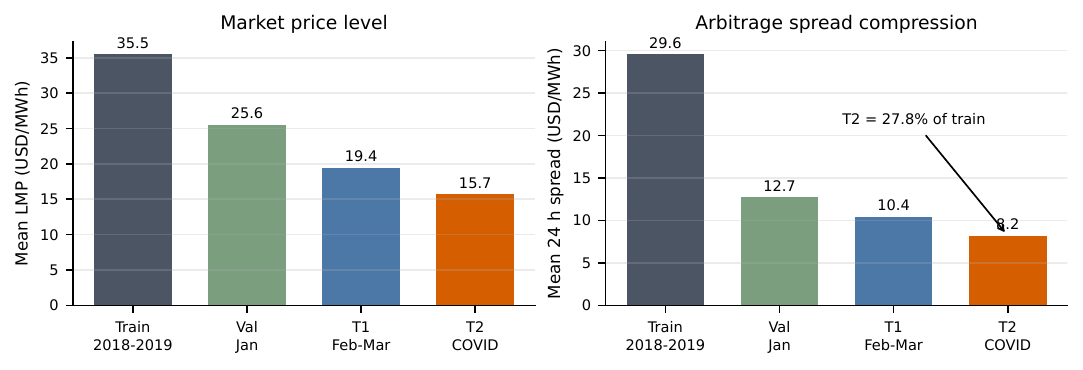}
\caption{Observed market context. T2 has both a lower mean price and a
compressed 24-hour arbitrage spread relative to the training period.}
\label{fig:market_regime}
\end{figure}

\subsection{Common-Information Controller Comparison}
Table~\ref{tab:controllers} applies one accounting contract and the same
causal forecast bundle to every forecast-dependent method.  PPO with advice
earns \$30.59 in T1 and
\$18.04 in T2, compared with
\$36.77 and
\$22.94 for proxy-cost
MPC.  The result places PPO below proxy-cost MPC under this protocol, with the
gap discussed against seed dispersion in
Section~\ref{sec:controller_interpretation}.  Price-only MPC and the rule controller cycle despite insufficient
net margin, showing that the economic specification materially affects the
comparison.

\begin{table}[t]
\centering
\caption{Common-accounting controller results. Values are mean net profit per
336-hour window (USD), T2 throughput is MWh, and $n$ is the number of controller
realizations. Deterministic baselines have $n=1$; PPO has five seeds.}
\label{tab:controllers}
\setlength{\tabcolsep}{3.2pt}
\begin{tabular}{lrrrr}
\toprule
Method & T1 net & T2 net & T2 MWh & $n$ \\
\midrule
Hold & 0.00 & 0.00 & 0.00 & 1 \\
Rule-based & -244.98 & -258.60 & 33.46 & 1 \\
Price MPC & -400.11 & -685.51 & 77.14 & 1 \\
Proxy-cost MPC & 36.77 & 22.94 & 2.71 & 1 \\
PPO, advice off & 30.59 & 16.45 & 2.36 & 5 \\
PPO, advice on & 30.59 & 18.04 & 2.18 & 5 \\
\bottomrule
\end{tabular}
\end{table}

Figure~\ref{fig:economic_robustness} exposes accounting and seed dispersion
without compressing positive results beside large cost-blind losses. For PPO
with advice, \$37.73 gross energy revenue plus \$2.12 clean-energy credit minus
\$21.80 throughput cost closes to the \$18.04 net objective. PPO has
5/5 non-negative T2 seed means; auxiliary NoVirus averages
\$9.71 with 1/5 negative seed means. This is
descriptive because training differs beyond one toggle.

\begin{figure*}[t]
\centering
\includegraphics[width=0.90\textwidth]{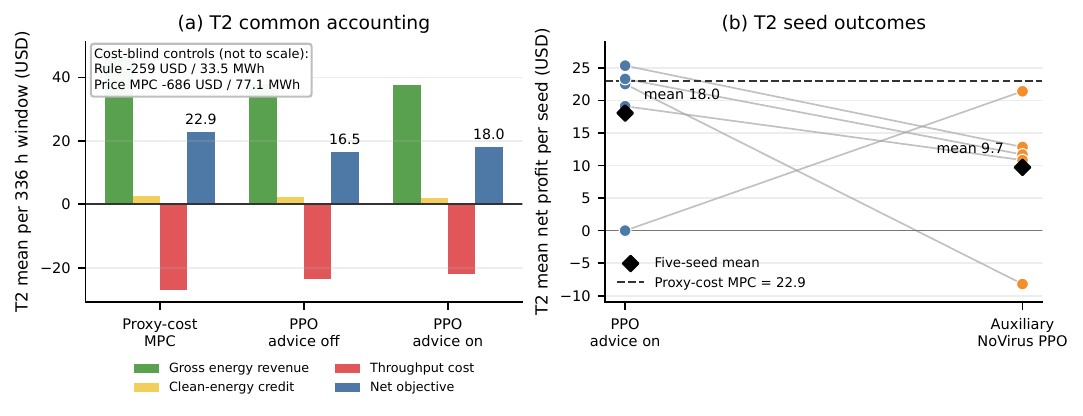}
\caption{T2 accounting and seed robustness. (a) Gross energy revenue,
clean-energy credit, negative throughput cost, and net objective; cost-blind
losses are annotated off-scale.
(b) Paired PPO and auxiliary NoVirus seeds; diamonds are five-seed means and
the dashed line is proxy-cost MPC. The training contrast is non-causal.}
\label{fig:economic_robustness}
\end{figure*}

\subsection{Causal Forecast Sensitivity}
Fixed checkpoints are tested on five T1 and eight T2 disjoint 168-hour blocks.
Among persistence, daily-naive, weekly-naive, and seasonal blend, the blend has lowest T2
price MAE (1.60 USD/MWh)
and spread MAE (1.87
USD/MWh). PPO with advice is stable across the three seasonal variants
(\$17.74--
\$18.12
per block); persistence yields
\$8.91, versus
\$22.74
and \$16.71
for proxy-cost MPC. Thus both controllers are forecast-sensitive, with a larger
decline for PPO, and MPC remains higher throughout. These disjoint 168-hour
blocks are a different window set from the possibly overlapping 336-hour main
evaluation; their absolute profit values are not directly comparable. No
causal profit--MAE relation is implied.

\subsection{Mechanism Separation}
The fixed-checkpoint M0--M6 ablation changes only mask, projection, or advice.
Table~\ref{tab:mechanisms} separates pre-selection filtering from correction. With the
pre-selection mask removed but projection retained (M2), the detector records
2080.4 T1 and
2258.6 T2 infeasible
requests per seed that reach projection. With both physical layers removed (M5), the observed mean ramp-violation
counts are 28.4 and
23.4. M0 records no thermal,
voltage, or ramp violations here; this is not a formal safety guarantee.

\begin{table*}[t]
\centering
\caption{Evaluation-time mechanism ablation (five-seed means). Net values are
USD per 336-hour window; final three columns are mean T2 counts per seed for
detected infeasible requests, projected actions, and observed ramp violations.}
\label{tab:mechanisms}
\setlength{\tabcolsep}{6pt}
\begin{tabular}{lrrrrrr}
\toprule
Variant & T1 net & T2 net & T2 MWh & Infeas. req. & Project & Ramp \\
\midrule
M0 Full & 30.59 & 18.04 & 2.18 & 0.0 & 0.0 & 0.0 \\
M1 No advice & 30.59 & 16.45 & 2.36 & 0.0 & 0.0 & 0.0 \\
M2 No mask & 29.58 & 17.28 & 2.20 & 2258.6 & 2258.6 & 0.0 \\
M3 No projection & 30.59 & 18.04 & 2.18 & 0.0 & 0.0 & 0.0 \\
M4 Physical only & 30.59 & 16.45 & 2.36 & 0.0 & 0.0 & 0.0 \\
M5 Advice only & 27.46 & 14.31 & 2.33 & 2257.4 & 0.0 & 23.4 \\
M6 Bare policy & 27.41 & 13.00 & 2.50 & 2410.6 & 0.0 & 23.2 \\
\bottomrule
\end{tabular}
\end{table*}

\subsection{Advisory Effect under Compressed Spreads}
Advice leaves T1 unchanged
(\$30.59 without/with) and increases T2
mean net profit from \$16.45 to
\$18.04 while reducing mean throughput
from 2.36 to
2.18\,MWh. The
134,400 mode-split records comprise 33,600 decisions in each of
two test splits, duplicated across the advice-on/off diagnostic modes; they are
not independent observations. Only $<5$, 5--10, and 10--20\,USD/MWh bins occur,
precluding a monotonic spread claim. Realised-margin labels are ex-post
diagnostics, never inputs.

\subsection{Conditional Effect of the Economic Advisory}
At each M0 state, advice-on continues the reference path while advice-off is
stepped once in a copy. Thus $\Delta$ in Fig.~\ref{fig:conditional_advisory}
is a same-state one-step contrast, not another rollout. Conditioning uses only
issued spread, current SoC, and the advice-off candidate.

In T2, all 185 changes among 33,600 M0 decision steps
occur in the 5--10 USD/MWh bin, and all 185 change the candidate to
idle. The advisory changes all 36 observed
charge candidates and 19.9\% of
710 discharge candidates as an
unweighted seed mean; the corresponding pooled count is
149/710
(21.0\%).
Among intervention steps, throughput falls by
-0.274 and
-0.331\,MWh
per step in the observed low- and mid-SoC cells. The associated immediate net
objective differences are mixed
(+0.49 and
-1.67 USD
per step), so the aggregate T2 improvement cannot be attributed to a uniformly
positive one-step effect. No high-SoC intervention state is observed; the
evidence therefore supports concentration in selected 5--10 USD/MWh,
low/mid-SoC conditions rather than a monotonic spread rule.

\begin{figure*}[t]
\centering
\includegraphics[width=0.90\textwidth]{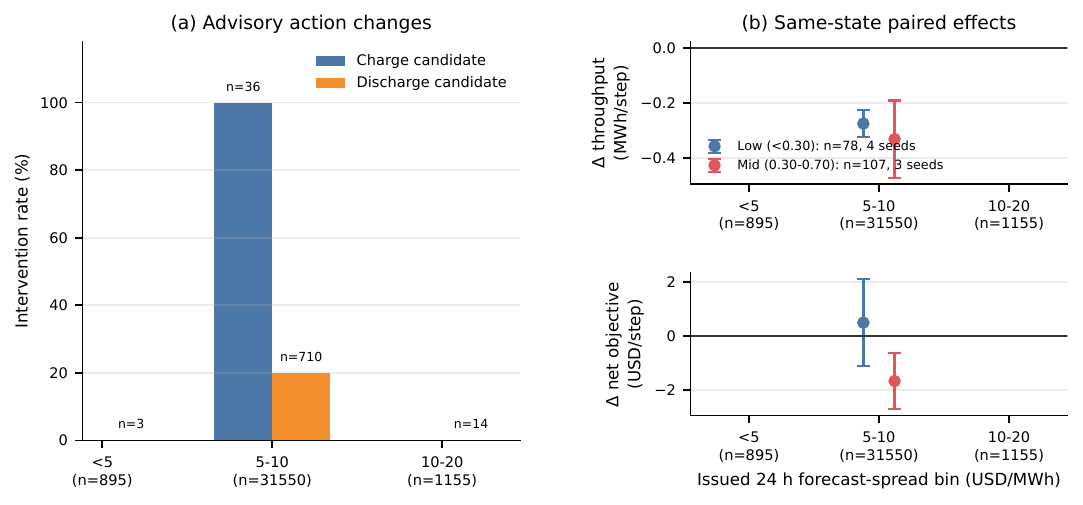}
\caption{T2 advisory diagnostic. (a) Advice-off candidate intervention rates;
blank denotes no observed candidate. (b) Advice-on minus advice-off one-step
effects from identical states; points are seed means and bars standard
deviations. Overlapping-window counts are descriptive; no high-SoC
interventions occur.}
\label{fig:conditional_advisory}
\end{figure*}

}{%
   \textbf{Camera-ready status:} corrected full reruns and the generated
   result artifact are required before submission.%
}

\section{Discussion}

\subsection{Mechanism-Level Interpretation}
The ablation separates three roles that aggregate reward or violation totals
cannot identify. The pre-selection mask prevents the policy from requesting actions outside
the implemented SoC, derating, and ramp constraints. Projection is a downstream
fallback: its high M2 count identifies infeasible requests that reach it after
mask removal; feasible execution does not imply that the policy's requests were
feasible. The identical M0 and M3
outcomes show that projection is not activated when the mask is present in
these test windows. Finally, M5's observed ramp violations establish why the
economic advisory cannot substitute for a physical layer. The conclusion is
limited to the implemented constraints and evaluated trajectories.

\subsection{Controller-Level Interpretation}
\label{sec:controller_interpretation}
Proxy-cost MPC exceeds PPO in both periods. In T2, the \$4.89 gap is smaller
than the PPO sample standard deviation across seed means (\$10.33).
Price-only MPC and rule losses appear after common throughput cost,
so cycling economics matters at least as much as optimizer choice. PPO has five
non-negative T2 seed means; auxiliary NoVirus has one negative seed, a
descriptive, non-causal contrast. PPO is also stable across three seasonal
forecasts but degrades under persistence, while proxy-cost MPC remains higher.
The dispersion contextualizes the gap without establishing equivalence, RL
dominance, curriculum causality, or a causal profit--MAE relation.

Offline training amortizes PPO policy optimization, but this implementation
shows neither a profit nor a runtime advantage over proxy-cost MPC (2.50 versus
1.50\,ms per decision); both times are negligible at hourly control. PPO is
therefore a test case for coupling a learned policy to explicit feasibility
layers under opportunity shift, not evidence that RL is preferable to MPC.

\subsection{Scope of the Economic Advisory}
Advice changes T2 but not T1. All observed T2 changes occur at 5--10 USD/MWh;
none occur in the observed adjacent bins. Paired effects are positive at low
SoC and negative at mid SoC, so the window gain is not a uniform one-step
benefit; no high-SoC interventions occur. Ex-post margin labels are diagnostic
only and create no look-ahead access.

\subsection{Limitations}
This is short-horizon dispatch, not lifecycle optimization: degradation is an
unvalidated throughput proxy and the reduced-order plant is uncalibrated.
Forecast sensitivity uses four deterministic constructions, not
probabilistic uncertainty; advisory evidence is limited to observed spreads
below 20 USD/MWh. Main 336-hour windows may overlap; disjoint 168-hour blocks
form a different window set from the same historical periods, so their absolute
profits are not comparable. PPO also operates at low T2 throughput (2.18\,MWh
per 336-hour window versus 2.36\,MWh without advice); whether an intermediate
economic setting increases trading while retaining the observed feasibility
outcomes is untested.

\section{Conclusion}

With identical causal information and accounting, five-seed PPO remains below
proxy-cost MPC. Advice raises only T2 profit while reducing throughput, with
changes concentrated at 5--10 USD/MWh and mixed SoC effects; PPO degrades under
persistence. The M0--M6 ablation separates economic screening from the two
physical layers: removing both exposes ramp violations, whereas protected runs
show none. The evidence does not establish RL superiority, curriculum efficacy,
formal safety, or lifecycle economics; those claims require calibrated plants,
probabilistic forecasts, external periods, and validated marginal-aging costs.

\bibliographystyle{IEEEtran}

\begin{thebibliography}{99}

\bibitem{mpc_ref}
E.~Perez, H.~Beltran, N.~Aparicio, and P.~Rodriguez,
``Predictive power control for PV plants with energy storage,''
\textit{IEEE Trans. Sustainable Energy}, vol.~4, no.~2,
pp. 482--490, 2013.

\bibitem{rl_bess_ref}
K.~Zhou, K.~Zhou, and S.~Yang,
``Reinforcement learning-based scheduling strategy for energy storage in
microgrid,''
\textit{J. Energy Storage}, vol.~51, Art. no.~104379, 2022.

\bibitem{perez2026}
J.~P\'erez, G.~Lobos, and M.~Bonacic,
``Integrated forecasting and deep reinforcement learning for price-based
self-scheduling of PV-BESS: Utility-scale evidence in Chile,''
\textit{PLoS ONE}, vol.~21, no.~1, Art. no.~e0336753, 2026.

\bibitem{ppo_ref}
J.~Schulman, F.~Wolski, P.~Dhariwal, A.~Radford, and O.~Klimov,
``Proximal policy optimization algorithms,''
\textit{arXiv preprint arXiv:1707.06347}, 2017.

\bibitem{sac_ref}
T.~Haarnoja, A.~Zhou, P.~Abbeel, and S.~Levine,
``Soft actor-critic: Off-policy maximum entropy deep
reinforcement learning with a stochastic actor,''
in \textit{Proc. ICML}, 2018, pp. 1861--1870.

\bibitem{milp_drl_ref}
G.~Xu \textit{et al.},
``An optimal solutions-guided deep reinforcement learning
approach for online energy storage control,''
\textit{Appl. Energy}, vol.~361, Art. no.~122915, 2024.

\bibitem{cao2020drl}
J.~Cao, D.~Harrold, Z.~Fan, T.~Morstyn, D.~Healey, and K.~Li,
``Deep reinforcement learning-based energy storage arbitrage with accurate
lithium-ion battery degradation model,''
\textit{IEEE Trans. Smart Grid}, vol.~11, no.~5, pp.~4513--4521, 2020.

\bibitem{invalid_action_masking}
S.~Huang and S.~Ontan\'on,
``A closer look at invalid action masking in policy gradient algorithms,''
in \textit{Proc. 35th Int. FLAIRS Conf.}, 2022.

\bibitem{chen2023shield}
P.~Chen, S.~Liu, X.~Wang, and I.~Kamwa,
``Physics-shielded multi-agent deep reinforcement learning for safe active
voltage control with photovoltaic/battery energy storage systems,''
\textit{IEEE Trans. Smart Grid}, vol.~14, no.~4, pp.~2656--2667, 2023.

\bibitem{gros2020projection}
S.~Gros, M.~Zanon, and A.~Bemporad,
``Safe reinforcement learning via projection on a safe set: How to achieve
optimality?'' \textit{IFAC-PapersOnLine}, vol.~53, no.~2,
pp.~8076--8081, 2020.

\bibitem{rasic2025safe_bess}
C.~Rasic, P.~Favaro, Y.~Wang, and J.-F.~Toubeau,
``Safe reinforcement learning for battery energy storage participation in the
imbalance settlement,'' \textit{IEEE Trans. Energy Markets, Policy Regul.},
vol.~4, no.~2, pp.~292--305, 2026.

\bibitem{fcrn_ref}
H.~Aaltonen, S.~Sierla, V.~Kyrki, M.~Pourakbari-Kasmaei, and V.~Vyatkin,
``Bidding a battery on electricity markets and minimizing
battery aging costs: A reinforcement learning approach,''
\textit{Energies}, vol.~15, no.~14, Art. no.~4960, 2022.

\bibitem{cpo_ref}
J.~Achiam, D.~Held, A.~Tamar, and P.~Abbeel,
``Constrained policy optimization,''
in \textit{Proc. 34th ICML}, PMLR, vol.~70, pp.~22--31, 2017.

\bibitem{nyiso2020}
Potomac Economics,
``2020 State of the Market Report for the New York ISO
Electricity Markets,''
NYISO, 2021.

\bibitem{drspcrl_ref}
A.~Satheesh, K.~Powell, and V.~Aggarwal,
``Distributionally robust self-paced curriculum reinforcement
learning,''
\textit{Reinf. Learn. J.}, accepted, 2026,
arXiv:2511.05694.

\bibitem{curriculum_ref}
Y.~Bengio, J.~Louradour, R.~Collobert, and J.~Weston,
``Curriculum learning,''
in \textit{Proc. ICML}, pp.~41--48, 2009.

\bibitem{liu2021}
X.~Liu, Z.~Zheng, \.{I}.~E.~B\"uy\"uktahtak\i n, Z.~Zhou, and P.~Wang,
``Battery asset management with cycle life prognosis,''
\textit{Reliab. Eng. Syst. Saf.}, vol.~216,
Art. no.~107948, 2021.

\bibitem{barletta2022}
G.~Barletta, P.~Di Prima, and D.~Papurello,
``Th\'evenin's battery model parameter estimation
based on Simulink,''
\textit{Energies}, vol.~15, no.~17, Art. no.~6207, 2022.

\bibitem{google_cfe}
Google,
``24/7 Carbon-Free Energy: Methodologies and Metrics,''
White Paper, Feb. 2021.

\end{thebibliography}

\end{document}